%% file: science_template.tex
\documentclass[12pt]{article}

\usepackage{newtxtext,newtxmath}

\usepackage{graphicx}

\usepackage[letterpaper,margin=1in]{geometry}

\renewenvironment{abstract}
	{\quotation}
	{\endquotation}

\date{}

\makeatletter
\renewcommand{\fnum@figure}{\textbf{Figure \thefigure}}
\renewcommand{\fnum@table}{\textbf{Table \thetable}}
\makeatother

\usepackage{scicite}

\usepackage{url}

\usepackage{lineno}
\usepackage{color}
\usepackage{comment}
\usepackage{subfigure}
\usepackage{graphicx}
\usepackage{multirow}
\usepackage{siunitx}
\usepackage{cite}
\usepackage{caption}
\usepackage{threeparttable}
\usepackage{tablefootnote}
\usepackage{algorithm, algorithmic}
\usepackage{xspace}
\usepackage{booktabs}
\usepackage{marvosym}
\usepackage{ifsym}
\usepackage{makecell}
\usepackage{newtxtext,newtxmath}
\usepackage{bm}
\usepackage{amsfonts}
\usepackage{colortbl}
\usepackage{xcolor}
\usepackage{tikz}
\usepackage{scicite}
\newcommand*{\circled}[1]{\lower.7ex\hbox{\tikz\draw (0pt, 0pt)%
    circle (.45em) node {\makebox[1em][c]{\small#1}};}}

\input{math_commands.tex}

\def\scititle{
        \centering
        STELLA: Self-Evolving LLM Agent for\\ Biomedical Research
}

\title{\bfseries \boldmath \scititle}

\author{
Ruofan Jin$^{1\ast}$,
Zaixi Zhang$^{1\ast\dagger}$,
Mengdi Wang$^{1}$,
Le Cong$^{2}$\\
\small $^{1}$Princeton University, Princeton, NJ, USA\\
\small $^{2}$Stanford University, Stanford, CA, USA\\
\small $^\ast$Co-first authors \quad
\small $^\dagger$Corresponding author: zz8680@princeton.edu
}

\begin{document} 

\maketitle

\begin{abstract} \bfseries \boldmath
The rapid growth of biomedical data, tools, and literature has created a fragmented research landscape that outpaces human expertise. While AI agents offer a solution, they typically rely on static, manually-curated toolsets, limiting their ability to adapt and scale. Here, we introduce STELLA, a self-evolving AI agent designed to overcome these limitations. STELLA employs a multi-agent architecture that autonomously improves its own capabilities through two core mechanisms: an evolving Template Library for reasoning strategies and a dynamic Tool Ocean that expands as a Tool Creation Agent automatically discovers and integrates new bioinformatics tools. This allows STELLA to learn from experience. We demonstrate that STELLA achieves state-of-the-art accuracy on a suite of biomedical benchmarks, scoring approximately 26\% on Humanity’s Last Exam: Biomedicine, 54\% on LAB-Bench: DBQA, and 63\% on LAB-Bench: LitQA, outperforming leading models by up to 6 percentage points. More importantly, we show that its performance systematically improves with experience; for instance, its accuracy on the Humanity’s Last Exam benchmark almost doubles with increased trials. STELLA represents a significant advance towards AI Agent systems that can learn and grow, dynamically scaling their expertise to accelerate the pace of biomedical discovery.
\end{abstract}


\section*{Introduction}

Modern biomedical research is defined by both immense opportunity and staggering complexity. As a cornerstone of science, it generates vast quantities of data from large-scale experiments, but this progress is hampered by a research landscape that is profoundly fragmented \cite{botvinik-nezer_variability_2020, thiele_protocol_2010, gibney_scientists_2013}. The knowledge, specialized software, and databases required to make discoveries are numerous, constantly evolving, and dispersed, forcing researchers to expend significant time and effort on the manual and labor-intensive task of discovering, learning, and integrating these disparate resources. While the advent of AI agents holds the promise of automating this intricate work \cite{wang_scientific_2023, tom_self-driving_2024, peng_study_2023}, current systems inherit a critical limitation: they typically rely on manually curated, static toolsets \cite{qu_crispr-gpt_2024, swanson_virtual_2024, roohani_biodiscoveryagent_2025, wang_txgemma_2025, gao2025txagent, xiao_cellagent_2024, zhang2025origene, huang2025biomni}. This approach is inefficient, fails to scale, and cannot keep pace with the rapid evolution of biomedical science, leaving the agents perpetually behind the cutting edge. This raises a critical question: Can we design a self-evolving agent that transcends these limitations by automatically discovering and integrating new tools, continuously updating its knowledge base, and iteratively upgrading its own capabilities through direct experience?

\begin{figure*}[t!]
    \centering
    \includegraphics[width=0.98\linewidth]{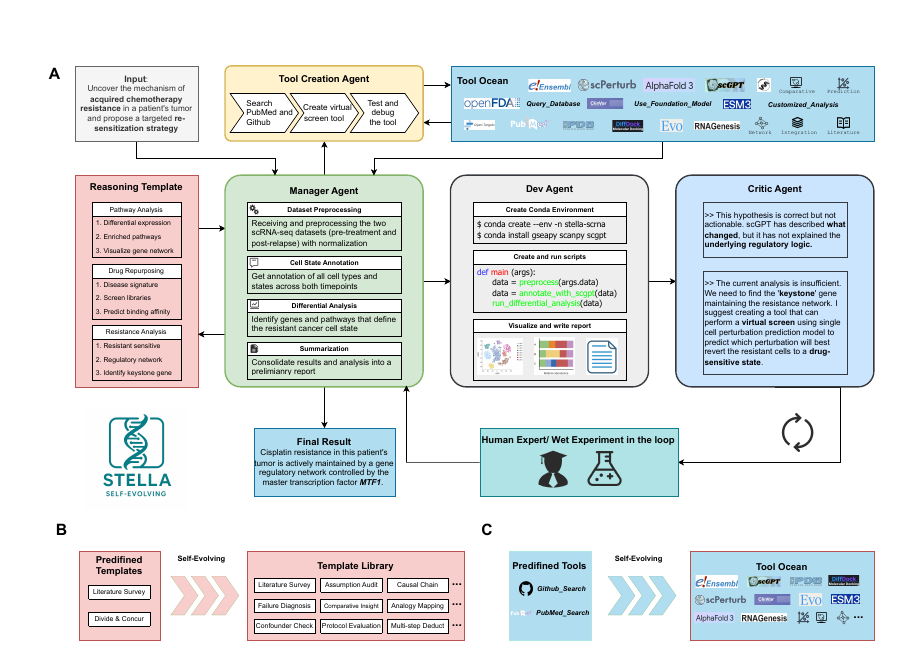}
    \caption{Overall Framework of STELLA, a self-evolving LLM Agent for Biomedical Research. (\textbf{A}) STELLA leverages four key agents, including Manager Agent, Dev Agent, Critic Agent, and Tool Creation Agent. The manager agent coordinates all agents and curates a reasoning template library to leverage successful reasoning experience; dev agent focuses on environment building, code creation, model training, and report writing; critic agents reflects on the intermediary results and provide suggestions; tool creation agent identifies the gap of agent capabilities and create new tools stored in Tool Ocean. Human expert and wet experiment results can provide valuable feedback and guidance in the loop. (\textbf{B} and \textbf{C}) Two key features of STELLA's self-evolving mechanisms. The Template Library evolves by including successful previous examples; the Tool Ocean evolves from simple predefined tools during agent inference. }   \label{illustration}
\end{figure*}

Here we present STELLA, a generalist biomedical AI agent designed around the core principle of \textbf{self-evolution} \cite{qiu2025alita}. STELLA learns and improves from every problem it solves, continuously enhancing its own reasoning strategies and technical abilities. Its architecture leverages four key agents---a Manager, Developer, Critic, and Tool Creation Agent---that work in concert to orchestrate complex tasks (Fig. \ref{illustration}A). Given a research prompt, the Manager Agent coordinates a multi-step reasoning plan. The Dev Agent then executes these steps by generating and running Python code to perform complex bioinformatics analyses. Throughout this process, the Critic Agent assesses intermediate results, identifying flaws and providing actionable feedback to refine the approach, creating a robust, iterative problem-solving loop \cite{wei_chain-of-thought_2022, yao_react_2023, wang_executable_2024}.

\begin{figure*}[t!]
    \centering
    \includegraphics[width=0.98\linewidth]{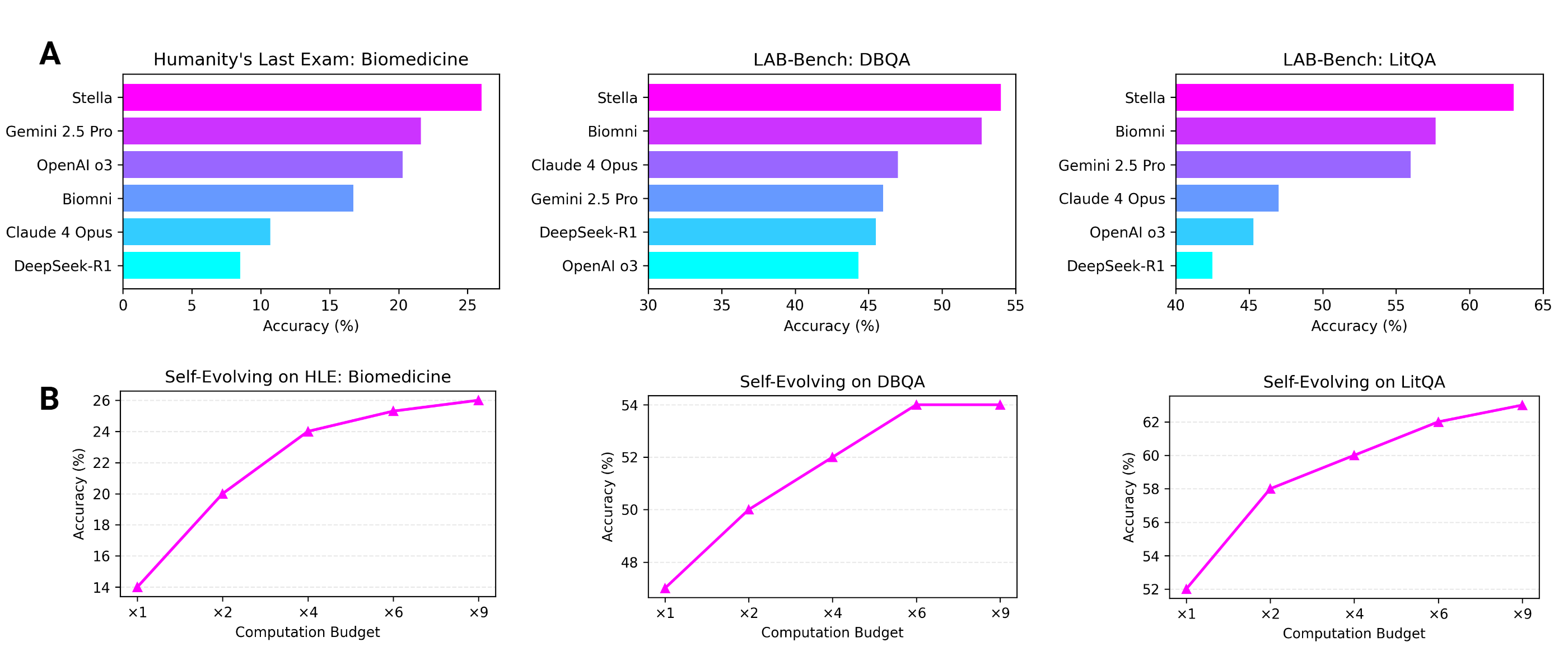}
    \caption{(\textbf{A}) Benchmark results of Stella with state-of-the-art LLMs and agents on Humanity's Last Exam: Biomedicine and LAB-Bench: DBQA and LitQA. (\textbf{B}) Test-time self-evolving effects on Benchmarks. The computation budget indicates the number of trials. The reported results represent the average accuracy across three independent evaluation runs.}
    \label{result}
\end{figure*}

The key to STELLA's advancement lies in its two novel self-evolving mechanisms. First, a \textbf{Template Library} of reasoning workflows is dynamically updated with successful strategies, allowing STELLA to learn from and generalize its problem-solving approaches over time (Fig. \ref{illustration}B). Second, its \textbf{Tool Ocean}, which contains STELLA's accessible bioinformatics tools, databases, and APIs, is not fixed. The Tool Creation Agent can autonomously identify, test, and integrate new tools in response to the demands of a novel problem, ensuring its capabilities are never limited to a predefined set (Fig. \ref{illustration}C). This integrated system allows STELLA to not only tackle challenging, large-scale biomedical problems with high efficiency but also to grow more capable with experience. We demonstrate that STELLA achieves state-of-the-art performance across a suite of demanding biomedical benchmarks \cite{phan_humanitys_2025, laurent_lab-bench_2024} (Fig. \ref{result}A) and, crucially, that its accuracy systematically improves with use, providing direct evidence of its self-evolving design (Fig. \ref{result}B). STELLA thus represents a significant advance towards AI systems that can learn and grow like a human scientist, dynamically scaling to meet the ever-expanding challenges of biomedical discovery.

Our empirical evaluations validate the efficacy of STELLA’s architecture. Across a diverse suite of challenging biomedical reasoning benchmarks, STELLA consistently establishes a new state of the art, achieving top accuracy scores of approximately 26\% on Humanity’s Last Exam: Biomedicine, 54\% on LAB-Bench: DBQA, and 63\% on LAB-Bench: LitQA, outperforming the next-best models by up to 8 percentage points (Fig. \ref{result}A). Critically, we provide direct evidence for its core self-evolving capability. This improvement is substantial; with increased computational experience, STELLA's accuracy on HLE: Biomedicine benchmark almost doubles, rising from 14\% to 26\% (Fig. \ref{result}B). This validates that STELLA not only performs at a superior level but also grows more capable with experience, effectively learning how to be a better scientist over time.

\section*{Results}
\subsection*{STELLA's Overall Framework}
STELLA leverages four key agents---a \textbf{Manager Agent}, \textbf{Dev Agent}, \textbf{Critic Agent}, and \textbf{Tool Creation Agent}---to systematically address complex biomedical research questions (Figure~\ref{result}A). The workflow begins when the \textbf{Manager Agent} receives a high-level research goal, such as to ``uncover the mechanism of acquired chemotherapy resistance and propose a re-sensitization strategy." The Manager Agent analyzes this goal and, guided by its reasoning experience, establishes a ``Reasoning Pathway''---a strategic plan that decomposes the problem into logical steps like `Differential expression analysis` and `Identify keystone gene`. It first assigns the initial data analysis tasks to the \textbf{Dev Agent}, which acts as a computational workhorse. The Dev Agent creates a self-contained conda environment and executes practical analysis scripts---for instance, running \texttt{diff\_analysis.py} to compare the transcriptomes of pre-treatment and post-relapse tumor samples.

The results of this initial analysis are then summarized and passed to the \textbf{Critic Agent} for rigorous evaluation. In the chemoresistance example, the Critic provides crucial feedback such as:
    \emph{This hypothesis is correct but not actionable... It has described \textbf{what} changed but not the underlying regulatory logic. We need to find the `keystone' gene.}
This feedback identifies a critical capability gap. In response, the Manager Agent tasks the \textbf{Tool Creation Agent} to close this gap. This agent searches existing resources and leverages a powerful collection of predefined models and tools called the \textbf{Tool Ocean}---which includes models like to build, test, and validate a new, more powerful tool, such as a virtual perturbation screening model based on virtual cell foundation model state \cite{adduri2025predicting}. By deploying this new tool, STELLA moves beyond simple description to prediction, ultimately identifying the transcription factor \texttt{MTF1} as the keystone regulator of the resistance network.

\subsection*{STELLA's Self-evolving Mechanisms for Biomedical Research}

A defining feature of STELLA is its dual self-evolving capability, which allows it to learn from experience and continuously expand its own abilities (Figure~\ref{illustration}B, C). The first mechanism is the evolution of its \textbf{Template Library}. The successful multi-step workflow used to identify \texttt{MTF1}---from initial descriptive analysis to the pivot towards a predictive virtual screen---is not discarded. It is distilled into a new, high-quality reasoning template and saved in the library. This process refines STELLA's strategic knowledge, allowing it to solve similar "mechanism of resistance" problems more efficiently in the future.

The second, more profound level of evolution is the expansion of the \textbf{Tool Ocean}, a dynamic and growing collection of STELLA's executable capabilities. This ocean contains a diverse array of computational tools that can be broadly classified into three main categories: (1) functions for querying established scientific databases, (2) interfaces for leveraging large-scale foundation models, and (3) customized analysis tools. The first category provides direct access to vital data sources like \texttt{PubMed} \cite{white2020pubmed}, \texttt{ClinVar} \cite{landrum2016clinvar}, and protein structures from \texttt{PDB} \cite{sussman1998protein}. The second allows STELLA to harness the power of state-of-the-art AI, including models like \texttt{AlphaFold~3} \cite{abramson2024accurate} for protein structure prediction, \texttt{scGPT} \cite{cui2024scgpt} for single-cell data interpretation, and \texttt{ESM3} \cite{hayes2025simulating} for protein language modeling. The third category consists of specialized and custom-built scripts for tasks like network analysis, and data integration.

Together, the evolution of Template Library and Tool Ocean empowers STELLA to tackle increasingly complex biomedical challenges with growing autonomy and scientific sophistication.

\subsection*{STELLA Outperforms State-of-the-art LLMs and Agents}
To evaluate its effectiveness, STELLA was benchmarked against a suite of state-of-the-art large language models and specialized agents on three challenging biomedical question-answering tasks. The results, presented in Figure~\ref{result}A, demonstrate that STELLA consistently achieves superior performance across all benchmarks. On the \texttt{Humanity's Last Exam (Biomedicine)} \cite{phan2025humanity} benchmark, STELLA attained a top accuracy of 26\%, surpassing all other tested models. This lead was extended on the \texttt{LAB-Bench} \cite{laurent_lab-bench_2024} suite, where STELLA achieved the highest scores of approximately 54\% on the DBQA task and 63\% on the LitQA task. These results validate the efficacy of its integrated multi-agent architecture compared to both generalist models like Gemini 2.5 Pro \cite{GoogleGeminiPro} and other specialized agents.

Furthermore, Figure~\ref{result}B highlights a core strength of the framework: its test-time self-evolving capability. These results show a clear and positive correlation between computational budget and performance. As the number of iterative trials increases from 1x to 9x, STELLA's accuracy exhibits a consistent and significant improvement across all three benchmarks. For instance, on the LitQA task, accuracy rises from approximately 52\% at a 1x budget to 63\% at a 9x budget. This demonstrates that STELLA's self-evolving design effectively leverages increased computation to refine its strategies, correct errors, and ultimately enhance the quality of its final answer.

\section*{Conclusion}
In this work, we addressed a fundamental limitation of current AI agents in biomedical research: their reliance on static, predefined capabilities in a field defined by constant evolution. We introduced STELLA, a generalist agent built on the principle of self-evolution. By integrating a multi-agent architecture with two novel mechanisms—an adaptive Template Library for reasoning and a dynamic Tool Ocean for capabilities—STELLA can learn from experience, continuously expanding its own knowledge and skills.

Our results demonstrate that this self-evolving design is not only effective but transformative. STELLA not only achieved state-of-the-art performance across multiple challenging biomedical benchmarks but, more importantly, showed significant and systematic improvement as it gained experience. This capacity to learn and grow moves beyond simple automation and represents a paradigm shift from AI as a static tool to AI as a dynamic scientific partner.

The development of STELLA marks a critical step towards creating truly autonomous AI scientists that can keep pace with the rapid rate of discovery. While challenges remain in bridging the gap between benchmark performance and real-world laboratory application, the ability of an agent to autonomously identify and master new tools lays the groundwork for systems that can explore novel scientific frontiers. Future work will focus on deploying STELLA in real-world research workflows and enhancing its collaboration with human scientists. Ultimately, self-evolving agents like STELLA have the potential to democratize expertise, unlock new avenues of inquiry, and fundamentally accelerate the engine of biomedical discovery.

\section*{Methods}
\subsection*{Baselines}

To evaluate STELLA's performance against existing methods on Humanity's Last Exam and the LAB-Bench datasets, a comprehensive set of baseline models was selected, categorized into two main groups:

\begin{itemize}
    \item[\textbullet] \textbf{LLMs:} We included Gemini 2.5 Pro \cite{GoogleGeminiPro}, Claude 4 Opus \cite{AnthropicClaude4}, DeepSeek-R1 \cite{guo2025deepseek}, and OpenAI o3 \cite{OpenAIo3} as representative state-of-the-art LLMs that offer strong general knowledge and reasoning capabilities. Gemini 2.5 Pro is known for its large context window and strong performance on complex tasks. Claude 4 Opus is a highly capable model recognized for its advanced reasoning across a wide range of benchmarks. DeepSeek-R1 is noted for its advanced language understanding and reasoning skills, while OpenAI o3 represents a powerful, state-of-the-art model from OpenAI.
    
    \item[\textbullet] \textbf{Biomedical Agents:} Biomni \cite{huang2025biomni} was chosen as a domain-specific baseline. As a powerful agent explicitly designed to automate and advance biomedical research across a wide range of subfields, it provides the most direct and relevant comparison to STELLA’s performance in this specialized domain.
\end{itemize}

For STELLA, we use Claude 4 Sonnet for the Dev Agent and Tool Creation Agent. Gemini 2.5 Pro is used for the Manager Agent and Critic Agent.

\subsection*{Q\&A Benchmarks}

For a direct and fair comparison with leading LLMs and agents, we followed the experimental settings of Biomni \cite{huang2025biomni} and OriGene \cite{zhang2025origene} with some modifications and applied to two main benchmark suites:

\begin{itemize}
    \item \textbf{LAB-Bench (DBQA \& LitQA):} \cite{phan2025humanity} The testing sets were created by using the same 12.5\% sampled subset of the complete Database Question-Answering (DBQA) and Literature Question-Answering (LitQA) sub-benchmarks. No development sets are used. This provides a cost-effective yet representative assessment of model performance. Our evaluation strictly followed the official LAB-Bench protocol, using multiple-choice answer options and allowing for abstention due to insufficient information. 

    \item \textbf{Humanity’s Last Exam (HLE):} \cite{laurent2024lab} We followed the sampling protocol from the Biomni study, evaluating STELLA on a selected set of 50 representative questions from the benchmark. This question set spans fourteen subdisciplines of Biology and Medicine, including Genetics, Molecular Biology, Computational Biology, and Bioinformatics. The evaluation was conducted  on the test set following the established protocol.
\end{itemize}

\bibliography{main} 
\bibliographystyle{sciencemag}

\end{document}

%% file: math_commands.tex
\usepackage{amsfonts, bm}

\def\eqref#1{equation~\ref{#1}}

\def\1{\bm{1}}

\DeclareMathAlphabet{\mathsfit}{\encodingdefault}{\sfdefault}{m}{sl}
\SetMathAlphabet{\mathsfit}{bold}{\encodingdefault}{\sfdefault}{bx}{n}

